\documentclass[runningheads]{llncs}
\usepackage{graphicx}
\usepackage{color}
\usepackage{array}
\usepackage{multirow}
\usepackage[numbers]{natbib}

\begin{document}
\title{Hybrid Text Feature Modeling for Disease Group Prediction using Unstructured Physician Notes\thanks{Submitted to ICCS 2020.}}
\titlerunning{Disease Group Prediction using Unstructured Physician Notes}
%
\author{Gokul S Krishnan\orcidID{0000-0002-1344-4722} \and
Sowmya Kamath S\orcidID{0000-0002-0888-7238}}
\authorrunning{Krishnan and Kamath}
%
\institute{Healthcare Analytics and Language Engineering (HALE) Lab,\\
	Department of Information Technology,\\
	National Institute of Technology Karnataka\\
	Surathkal, INDIA - 575025\\
\email{gsk1692@gmail.com, sowmyakamath@nitk.edu.in}\\}
\maketitle              
\begin{abstract}
Existing Clinical Decision Support Systems (CDSSs) largely depend on the availability of structured patient data and Electronic Health Records (EHRs) to aid caregivers. However, in case of hospitals in developing countries, structured patient data formats are not widely adopted, where medical professionals still rely on clinical notes in the form of unstructured text. Such unstructured clinical notes recorded by medical personnel can also be a potential source of rich patient-specific information which can be leveraged to build CDSSs, even for hospitals in developing countries. If such unstructured clinical text can be used, the manual and time-consuming process of EHR generation will no longer be required, with huge person-hours and cost savings. In this paper, we propose a generic ICD9 disease group prediction CDSS built on unstructured physician notes modeled using hybrid word embeddings. These word embeddings are used to train a deep neural network for effectively predicting ICD9 disease groups. Experimental evaluation showed that the proposed approach outperformed the state-of-the-art disease group prediction model built on structured EHRs by 15\% in terms of AUROC and 40\% in terms of AUPRC, thus proving our hypothesis and eliminating dependency on availability of structured patient data.
\end{abstract}

\newcolumntype{P}[1]{>{\centering\arraybackslash}p{#1}} 
\newcolumntype{R}[1]{>{\flushright\arraybackslash}p{#1}} 
\newcolumntype{L}[1]{>{\flushleft\arraybackslash}p{#1}} 
\newcolumntype{M}[1]{>{\flushleft\arraybackslash}m{#1}}

\section{Introduction}
Electronic Health Records (EHRs) and predictive analytics have paved the way for Clinical Decision Support Systems (CDSSs) development towards realizing intelligent healthcare systems. The huge amount of patient data generated in hospitals in the form of discharge summaries, clinical notes, diagnostic scans etc, present a wide array of opportunities to researchers and data scientists for developing effective methods to improve patient care. CDSSs built on techniques like data mining and machine learning (ML) have been prevalent in healthcare systems, aiding medical personnel to make informed, rational decisions and execute timely interventions for managing critical patients. Some existing ML based CDSSs include -- mortality prediction \cite{krishnan2019novel,shickel2019deepsofa,ge2018interpretable,purushotham2018benchmarking,harutyunyan2017multitask,calvert2016using}, hospital readmission prediction \cite{jiang2018integrated,reddy2018predicting,kansagara2011risk}, length of stay prediction \cite{harutyunyan2017multitask,van2007optimizing,appelros2007prediction}, generic disease or ICD9\footnote{ICD-9-CM: International Classification of Diseases, Ninth Revision, Clinical Modification} group prediction \cite{purushotham2018benchmarking,miotto2016deep,choi2016doctor} and so on. An important point to note is that, most existing CDSSs depend on structured patient data in the form of Electronic Health Records (EHRs), which favors developed countries due to large scale adoption of EHRs. However, healthcare personnel in developing countries continue to rely on clinical notes in the form of free and unstructured text as EHR adoption is extremely low. Moreover, the conversion of patient information recorded in the form of text/clinical reports to a standard structured EHR is a labor-intensive, manual process that can result in loss of critical patient-specific information that might be present in the clinical notes. For this reason, there is a need for alternative approaches of developing CDSSs that do not rely on structured EHRs. 

ICD9 disease coding is an important task in a hospital, during which a trained medical coder with domain knowledge assigns disease-specific, standardized codes called ICD9 codes to a patient's admission record. As hospital billing and insurance claims are based on the assigned ICD9 codes, the coding task requires high precision, but is often prone to human error, which has resulted in an annual spending of \$25 billion in the US to improve coding efficacy \cite{xie2018neural,farkas2008automatic}. Hence, automated disease coding approaches have been developed as a solution to this problem. Currently, this is an area of active research \cite{zeng2019automatic,li2018automated,xie2018neural,baumel2018multi,berndorfer2017automated,ayyar2016tagging}, but performance that has been recorded so far is below par indicating huge scope for improvement, thus making automated ICD9 coding an open research problem. {\color{black}Existing ICD9 coding methods make use of only discharge summaries for coding, similar to the process adopted by human medical coders. However, as the records are digitized, other textual records under the same admission identifier might be available, which may provide additional patient-specific insights pertaining to the diagnosed diseases. Further, for ICD9 disease coding to be effective, the correct determination of generic disease categories or groups prior to specific coding is very crucial as information regarding generic disease groups can be informative for specific ICD9 coding. With this in mind, we explore other textual sources of patient data, utilizing physician notes to predict the generic ICD9 disease groups for the patient, following which actual ICD9 disease code prediction can be achieved. In this paper, we focus on effective prediction of ICD9 disease groups and the code prediction will be considered in future work.}

\begin{figure}[tp]
	\centering
	\includegraphics[width=\textwidth]{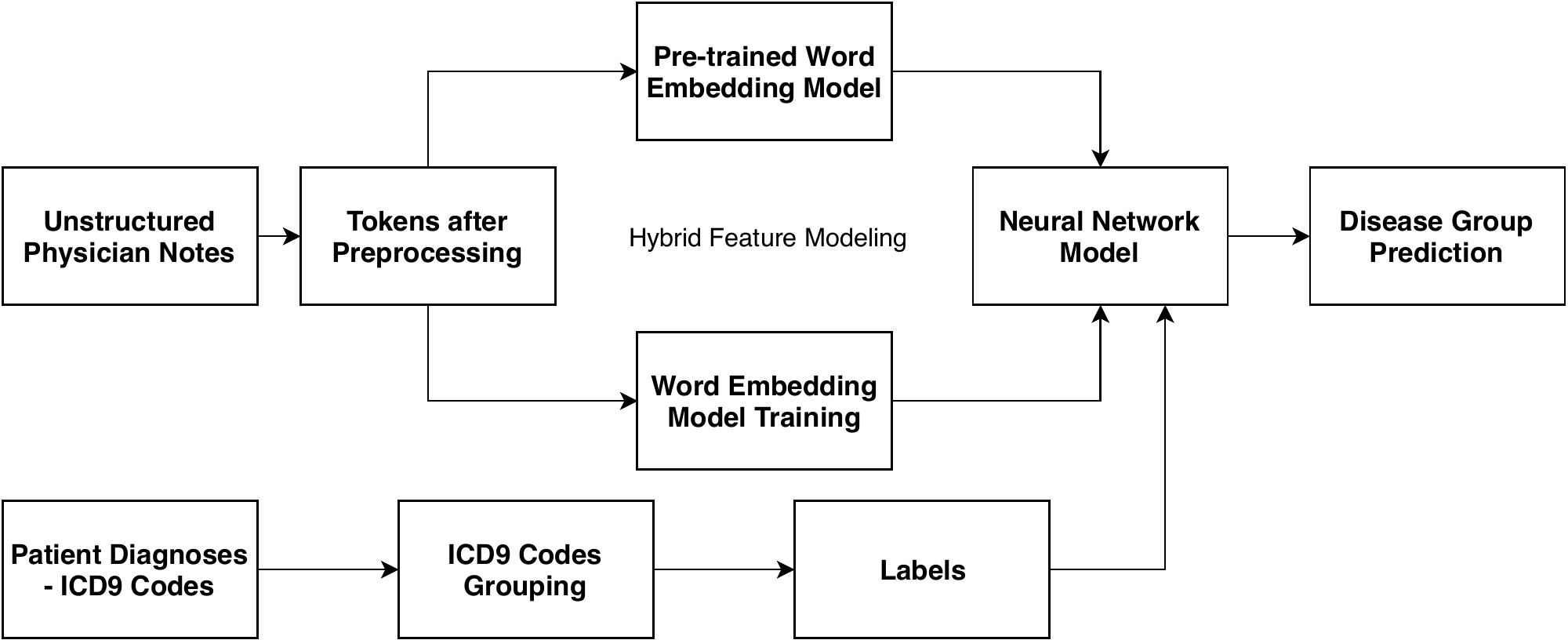}
	\caption{Proposed ICD9 Disease Group Prediction Process}
	\label{flow}
\end{figure}

{\color{black}While most existing works aim to predict ICD9 codes rather than ICD9 disease groups, \citet{purushotham2018benchmarking} performed benchmarking of three CDSS tasks - ICU mortality prediction, ICD9 group prediction and hospital readmission prediction on ICU patients data using Super Learner and Deep Learning techniques.} The ICD9 group prediction task discussed in their work is a generic disease group prediction task that is a step prior to ICD9 coding, which can also be a viable disease risk prediction CDSS for caregivers. Their work uses feature subsets extracted from structured patient data, where the best performing model uses multitude of numerical features such as input events (volume of input fluids through IV), output events (quantity of output events like urinary and rectal), prescriptions and other ICU events like chart events (readings recorded in the ICU) and lab events (lab test values). As deep learning models can effectively learn from numerous raw feature values quite well, their approach achieved good results. However, in a real world scenario, to measure all these values and then using them as inputs to a prediction model for an outcome, will cause an inevitable and significant time delay, during which the patient condition can worsen. CDSS prediction models that use unstructured clinical text data can predict desired outcomes with lower latency and also help improve the prediction performance as these raw text sources contain more patient-specific information. 

In this paper, a hybrid feature modeling approach that uses hybrid clinical word embeddings to generate quality features which are used to train and build a deep neural network model to predict ICD9 disease groups is presented. We show the results of a benchmarking study of our proposed model (modeled on unstructured physician notes) against the current state-of-the-art model for disease group prediction (built on structured clinical data). The rest of this paper is structured as follows: Section \ref{prop} describes the proposed approach in detail, followed by experimental results and discussion in Section \ref{results}. We conclude the paper with prospective future work and research directions. 

\section{Materials \& Methods}
\label{prop}
The proposed approach for disease group prediction is depicted in Fig. \ref{flow}. We used physicians' clinical notes in unstructured text form from the MIMIC-III \cite{johnson2016mimic} dataset for our experiments. MIMIC-III contains data relating to 63,000 ICU admissions of more than 46,000 patients admitted in Beth Israel Hospital, New York, USA, during 2001--2012. We extracted only the physician notes from the `NOTEEVENTS' table, which resulted in a total of 141,624 physician notes generated during 8,983 admissions. {\color{black}As per MIMIC documentation, physicians have reported some identified errors in notes present in the `NOTEEVENTS' table. As these notes can affect the training negatively, records with physician identified errors were removed from the cohort.} Additionally, those records with less than 15 words are removed and finally, the remaining 141,209 records are considered for the study. Some characteristic features of the physician notes corpus are tabulated in Table \ref{stat_corpus}. 

We observed that there were 832 kinds of physician notes available in the MIMIC-III dataset such as Physician Resident Progress note, Intensivist note, etc. The frequency statistics of the top ten kinds of physician notes are tabulated in Table \ref{phy_types}. A particular patient may suffer from multiple diagnoses during a particular admission and hence, it is necessary for the prediction to be a multi-label prediction task. Therefore, for each physician note, all the diagnosed disease groups during that particular admission were considered as labels and given binary values - 0 (if the disease was not diagnosed) and 1 (if the disease was diagnosed).

\begin{table}[h!]
	\centering
	\caption{Physician Notes Corpus Characteristics}
	\tabularnewline
	\centering
	\label{stat_corpus}
	\setlength\tabcolsep{8pt}
	\renewcommand{\arraystretch}{1.2}
	\begin{tabular}{lr}
		\hline
		\textbf{Feature}  & \textbf{Total Records} 
		\\ 
		\hline
		Physician Notes & 141,209   \\
		Unique Words     & 635,531   \\
		Words in longest Note & 3,443 \\
		Words in shortest Note & 16 	\\
		Average word length of notes & 858     \\
		Unique Diseases   & 4,208    \\
		Disease Groups    & 20     \\
		\hline
	\end{tabular}
\end{table}  

\begin{table}[h!]
	\centering
	\caption{Top 10 Physician Notes Types}
	\label{phy_types}
	\setlength\tabcolsep{8pt}
	\renewcommand{\arraystretch}{1.2}
	\begin{tabular}{M{6cm}r}
		\hline
		\textbf{Note Type} & \textbf{Occurrences} \\
		\hline
		Physician Resident Progress & 62,550 \\
		Intensivist & 26,028 \\
		Physician Attending Progress & 20,997 \\
		Physician Resident Admission & 10,611 \\
		ICU Note - CVI & 4,481 \\
		Physician Attending Admission (MICU) & 3,307 \\
		Physician Resident/Attending Progress (MICU) & 1,519 \\
		Physician Surgical Admission & 1,102 \\
		Physician Fellow/Attending Progress (MICU) & 970 \\
		Physician Attending Admission & 873 \\
		\hline
	\end{tabular}
\end{table}

\subsection{Preprocessing and Feature Modeling}The physician notes corpus is first preprocessed using basic Natural Language Processing (NLP) techniques such as tokenization and stop word removal. Using tokenization, the clinical text corpus is broken down into basic units called tokens and by the stopping process, unimportant words are filtered out. The preprocessed tokens are then fed into a pre-trained word embedding model to generate the word embedding vector representation of the corpus that can be considered as textual features. The pre-trained word embedding model used in this study is an openly available model that is trained on biomedical articles in PubMed and PMC along with texts extracted from an English Wikipedia dump \cite{nedellec2013overview}, hence capturing relevant terms and concepts in the biomedical domain, which helps generate quality feature representation of the underlying corpus. The pre-trained model generates word embedding vectors of size 1x200 for each word and these vectors were averaged to generate a representation such that each preprocessed physician note is represented as a 1x200 vector. The preproccesed tokens are then used to train a Word2Vec model \cite{mikolov2013efficient}, a neural network based word representation model that generates word embeddings based on co-occurrence of words. The Skipgram model of Word2Vec was used for training the physician notes tokens with a dimension size of 200 (same as the pre-trained model) with an initial learning rate of 0.025. The averaged word vector representation for each report tokens are extracted and then fed into the neural network model along with the vector representation extracted from the pre-trained model and the ICD9 disease group labels.


\subsection{ICD9 Disease Code Grouping}ICU Patients' diagnoses in terms of ICD9 disease codes from the `DIAGNOSES\_ICD' table of MIMIC-III dataset were grouped as per standards\footnote{Available online \url{http://tdrdata.com/ipd/ipd\_SearchForICD9CodesAndDescriptions.aspx}} (similar to the approach adopted by \citet{purushotham2018benchmarking}). A total of 4,208 unique ICD9 disease codes thus obtained were grouped into 20 ICD9 disease groups, i.e., potential labels. As the ICD9 group prediction task is considered as a binary classification of multiple labels, 20 labels (disease groups) were considered with binary values: 0 (negative diagnosis of the disease) and 1 (positive diagnosis of the disease). The physician notes modeled into two feature matrices of shape 141209 x 200 each, along with 20 ICD9 disease groups (labels) is now used to train the neural network model.

\begin{figure}[h!]
	\centering
	\includegraphics[scale=0.45]{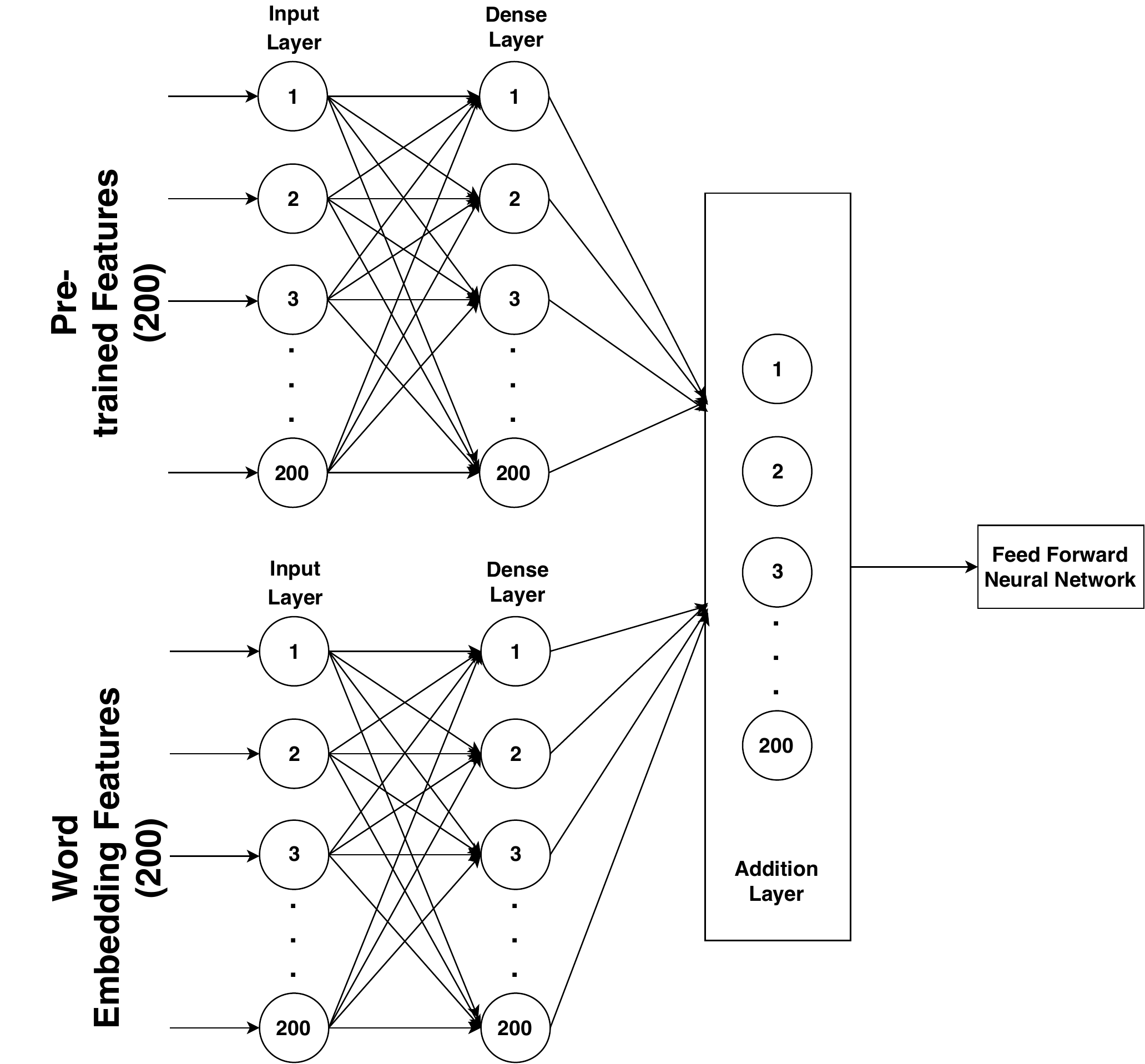}
	\caption{Hybrid Features using Dynamic Weighted Addition of Feature Representations}
	\label{input_add}
\end{figure}

\begin{figure}[ht!]
	\centering
	\includegraphics[scale=0.52]{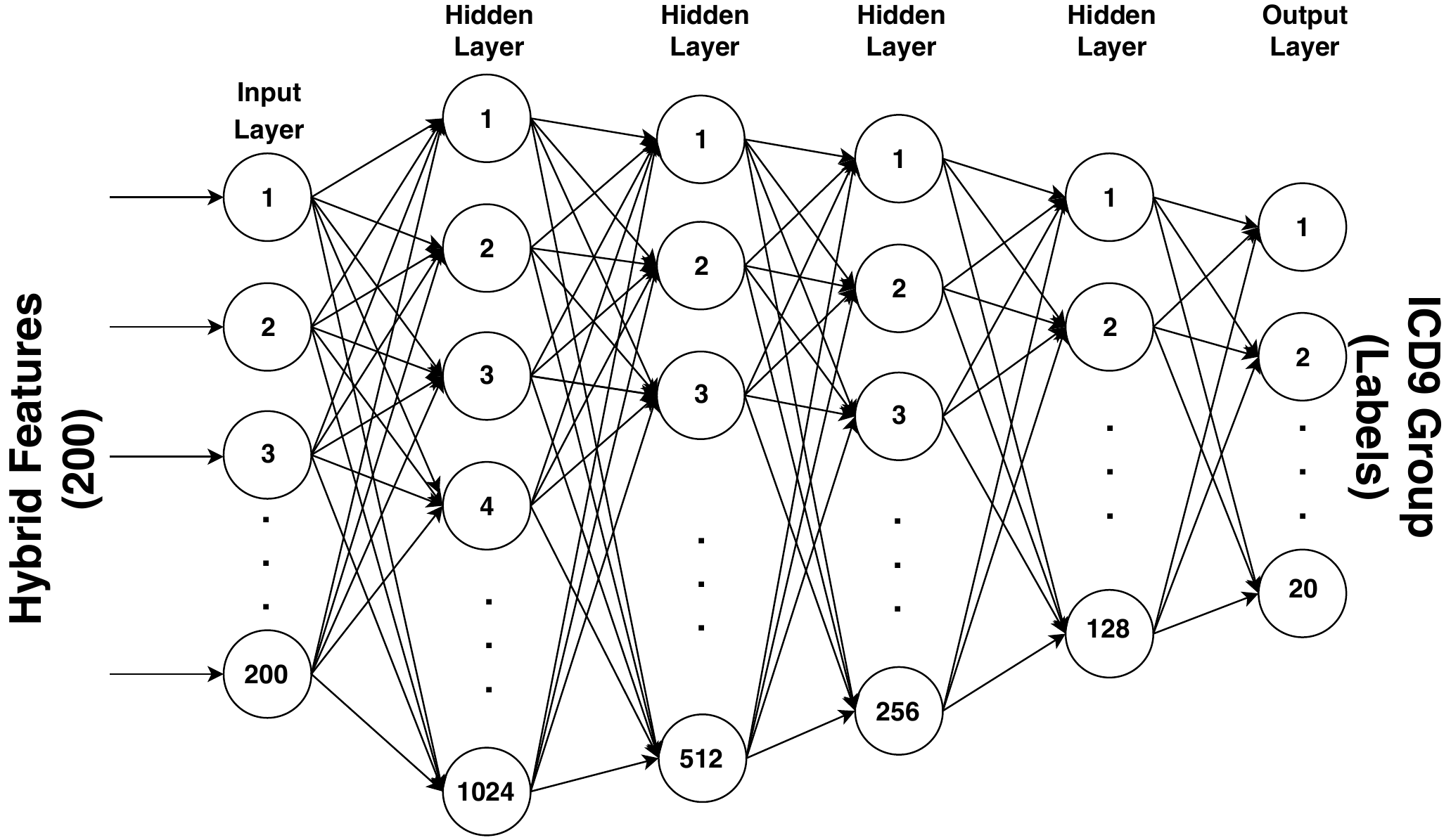}
	\caption{Feed-forward Neural Network - Disease Group Prediction Model}
	\label{model_nn}
\end{figure}

\subsection{Neural Network Model}
The proposed Deep Neural Network Prediction Model is illustrated in Fig. \ref{ffnn_model}. The neural network architecture is divided into 2 parts -- the first for determining the weights for the hybrid combination of features dynamically and the next for multi-label classification of ICD9 disease codes. 
%
%
The process of dynamically modeling the weightage to be assigned for the combination of pre-trained word embeddings and the word embeddings generated using the physicians notes is performed as shown in Fig. \ref{input_add}. The two feature set are fed as inputs into the neural network model, where both the input layers consists of 200 neurons, equal to the number of features generated from both models. The addition layer merges the two sets of input features in a weighted combination which is dynamically determined through backpropagation of the overall neural network architecture thereby ensuring the optimal and effective combination that offers the best classification performance possible. This architecture also ensures that the weights for the hybrid combination of features is always determined dynamically and hence can be used for any clinical text corpus. The combined set of features, i.e., the hybrid features, are then fed on to the dense feed forward neural network model which performs the training for multi-label classification of ICD9 disease code groups. 

\subsection{Disease Prediction Model}
\begin{figure}[ht!]
	\centering
	\includegraphics[scale=0.45]{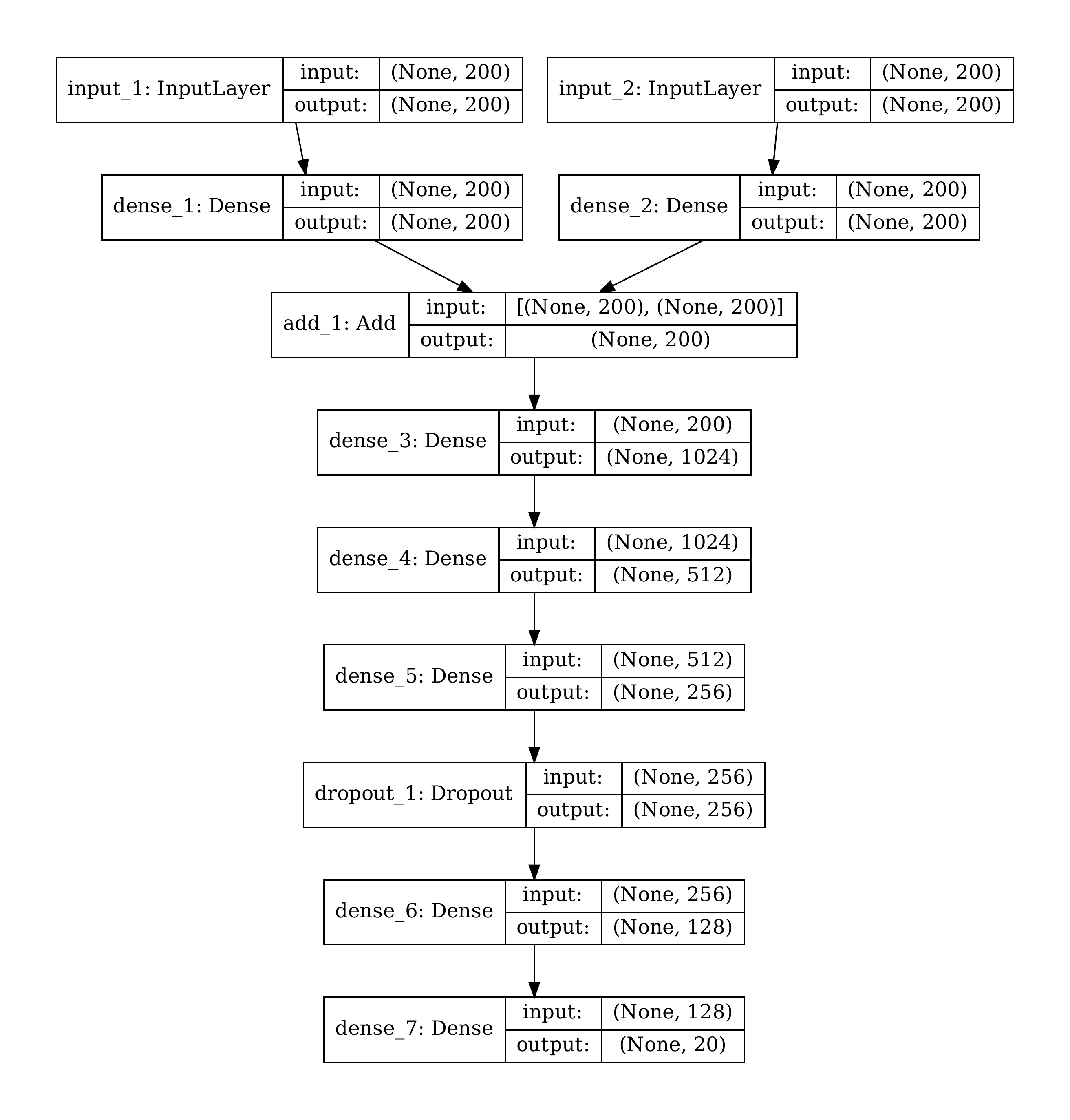}
	\caption{Overall Neural Network Model for ICD9 Group Prediction}
	\label{ffnn_model}
\end{figure}

The dynamically weighted feature matrix consisting of the hybrid word embedding features, along with ICD9 group labels are next used for training a Feed Forward Neural Network (FFNN) used as the prediction model (depicted in Fig. \ref{model_nn}). The input layer consists of 1024 neurons with input dimension as 200 (number of input features); followed by three hidden layers with 512, 256 and 128 neurons respectively and finally an output layer with 20 neurons, each representing an ICD-9 disease group. To prevent overfitting, a dropout layer, with a dropout rate of 20\% was also added to the FFNN model (see Figure \ref{ffnn_model}). As this is a binary classification for multiple labels, binary cross entropy was used as a loss function,  while Stochastic Gradient Descent (SGD) was used as the optimizer with a learning rate of 0.01. Rectified Linear Unit \textit{ReLU} activation function was used as the input and hidden layer activation functions as the feature matrix values are standardized to the range -1 and 1. The major hyperparameters for the FFNN model -- the optimizer, learning rate of the optimizer and the activation function, were tuned empirically over several experiments using the GridSearchCV function in Python sklearn library. Finally, the output layer activation function is a sigmoid, again as the classification is two-class for each of the 20 labels. Training was performed for 50 epochs and then the model was applied to the validation set to predict disease groups after which the results were observed and analyzed.

\section{Experimental Results \& Discussion}
\label{results}		

To evaluate the proposed approach, we performed several experiments using standard metrics like accuracy, precision, recall, F-score, Area under Receiver Operating Characteristic curve (AUROC), Area under Precision Recall Curve (AUPRC) and Matthew's Correlation Coefficient (MCC). 
We measured these metrics on a sample-wise basis, i.e., for each report, the predicted and actual disease groups were compared and analyzed. It can be observed from the Table 3 that, the proposed model achieved promising results: AUPRC of 0.85 and AUROC of 0.89. The accuracy, precision, MCC and F-score of 0.79, 0.82, 0.57 and 0.79 respectively also indicate a good performance. 

{\color{black}For evaluating the proposed hybrid text feature modeling approach, we designed experiments to compare its performance against that of baseline feature modeling approaches -- TF-IDF based bag-of-words approach, trained word embedding approach (only Word2Vec model) and a pre-trained word embedding approach (using word embedding model trained on PubMed, PMC and Wikipedia English articles). The bag-of-words approach is based on calculated term Frequency and inverse document frequency $(tfXidf)$ scores determined from the physician notes corpus. The Sklearn English stopword list was used to filter the stopwords and $n$-gram ($n=1,2,3$) features were considered. Finally, the top 1000 features were extracted from the corpus and then fed into the neural network model for training. The other two baselines were kept the same as explained in Section 2. It is to be noted that in the neural network configuration, only one input is present, i.e., there is no hybrid weighted addition layer. The results of comparison are tabulated in Table \ref{results_compare}. It can be observed that the proposed approach that involves a hybrid weighted combination of pre-trained and trained word embeddings is able to perform comparatively better in terms of all metrics.}

\begin{table}[h!]
	\caption{Experimental Results --  Baseline Comparison}
	\tabularnewline
	\label{results_compare}
	\centering
	\renewcommand*{\arraystretch}{1.1}	
	\setlength\tabcolsep{8pt}
	\begin{tabular}{lP{2cm}P{2.5cm}P{2cm}P{2cm}}			
		\hline
		\textbf{Parameter}                                                   & \textbf{Proposed Approach}                                                       & 
		\textbf{Bag-of-words (TF-IDF)}	& \textbf{Only Word2Vec}  & \textbf{Only Pre-trained}                                                \\ \hline
		AUROC & 0.89 & 0.87 & 0.88 & 0.88  \\
		AUPRC & 0.85 & 0.81 & 0.84 & 0.82  \\
		Accuracy & 0.79 & 0.78 & 0.79 & 0.79  \\
		Precision & 0.82 & 0.80 & 0.82 & 0.80 \\
		Recall & 0.79 & 0.78 & 0.79 & 0.78 \\
		F-Score & 0.79 & 0.78 & 0.79 & 0.79 \\
		MCC & 0.58 & 0.53 & 0.57 & 0.56 \\ 
		\hline
	\end{tabular}
\end{table}

\begin{figure}[h!]
	\centering
	\includegraphics[scale=0.3]{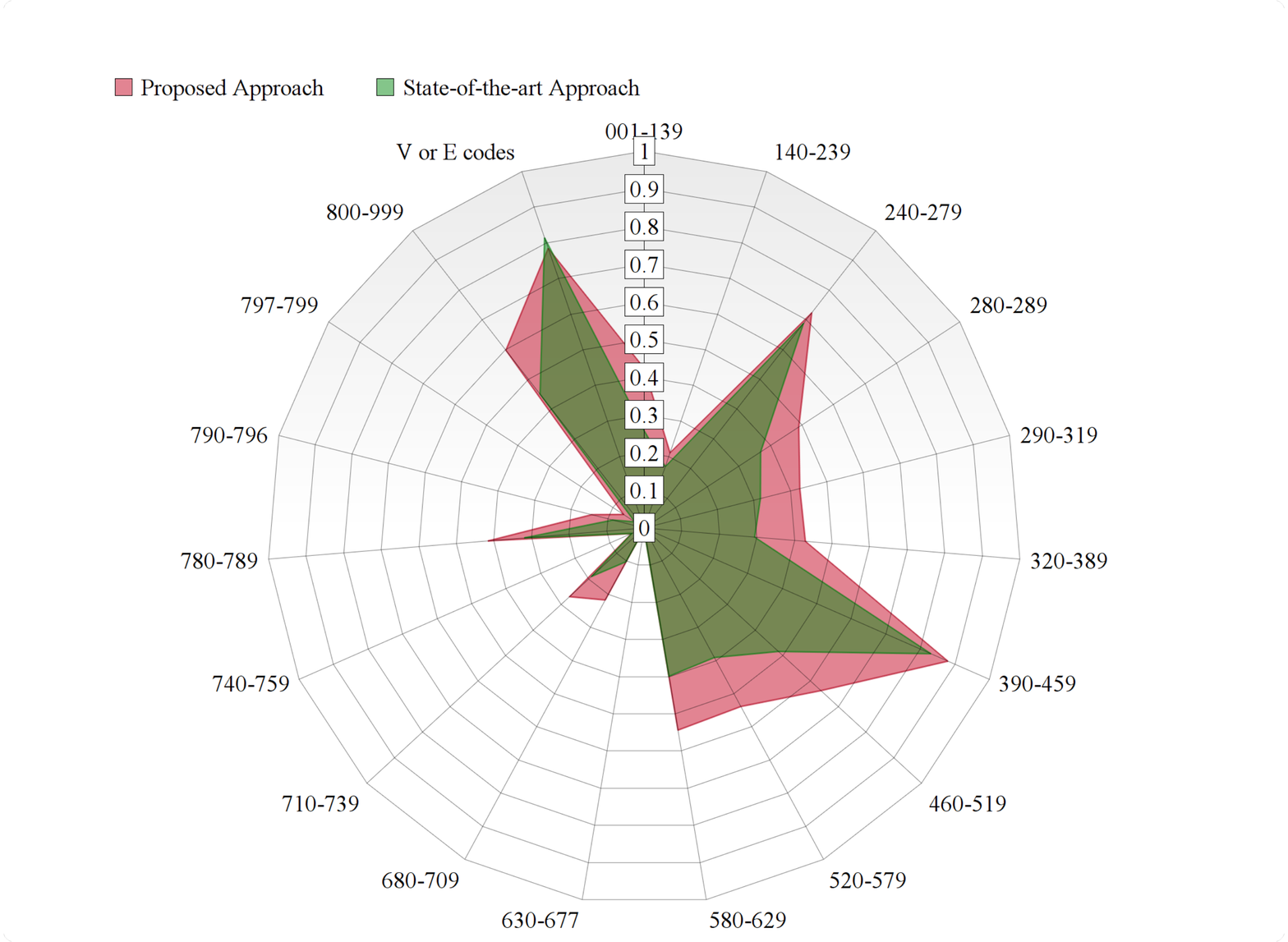}
	\caption{ICD9 Group Labels Statistics Comparison}
	\label{labels_ratio}
\end{figure}

\begin{table}[h!]
	\caption{Experimental Results -- Comparison with State-of-the-art}
	\tabularnewline
	\label{tab_results}
	\centering
	\renewcommand*{\arraystretch}{1.1}	
	\setlength\tabcolsep{8pt}
	\begin{tabular}{lcc}			
		\hline
		\textbf{Parameter}                                                   & \textbf{Our Approach}                                                       & \citet{purushotham2018benchmarking}                                                 \\ \hline
		Type of Data & Unstructured text                                                                & Structured data                                                             \\
		AUROC                                                             & 0.89                                                                     & 0.77                                                              \\
		AUPRC                                                             & 0.85                                                                   & 0.60                                             \\
		Accuracy                                                          & 0.79                                                                             & *                                                                           \\
		Precision                                                         & 0.82                                                                             & *                                                                           \\
		Recall                                                            & 0.79                                                                             & *                                                                           \\
		F-Score                                                           & 0.79                                                                             & *                                                                           \\
		MCC                                                               & 0.58                                                                             & *                                                                           \\ \hline
		\multicolumn{3}{l}{* Results not reported in the study}
	\end{tabular}
\end{table}

{\color{black}Next, a comparative benchmarking of the proposed approach against the state-of-the-art ICD9 disease group prediction model developed by \citet{purushotham2018benchmarking} was performed. For all the hospital admission IDs (HADM\_ID) included in the state-of-the-art approach \cite{purushotham2018benchmarking}, the physician notes, if present were extracted from `NOTEVENTS' table. Although the number of records under consideration for both the studies are different, it is to be noted that the labels are distributed similarly (statistics shown in Fig. \ref{labels_ratio}) and is therefore considered a fair comparison.
We consider this comparison in order to study the effect of the disease group prediction models built on structured patient data (state-of-the-art approach \cite{purushotham2018benchmarking}) and unstructured patient data (physician notes in this case)}. The results of the benchmarking are tabulated in Table \ref{tab_results}, which clearly shows the proposed approach outperformed Purushotham et al's model \citet{purushotham2018benchmarking} by 15\% in terms of AUROC and 40\% in terms of AUPRC. This shows that the predictive power of a model built on unstructured patient data exceeds that of those built on structured data, where some relevant information may be lost during the coding process. To encourage comparative studies that use physician notes in MIMIC-III dataset, certain additional metrics were also considered. The Recall \& F-Score performance as well as the MCC values of the proposed model over our easily reproducible patient cohort data subset are also observed and provided. 



\textbf{Dicussion}:
From our experiments, we observed a significant potential in developing prediction based CDSS using unstructured text reports directly, eliminating the dependency on the availability of structured patient data and EHRs. The proposed approach that involves a textual feature modeling and a neural network based prediction model was successful in capturing the rich and latent clinical information available in unstructured physician notes, and using it to effectively learn disease group characteristics for prediction. The Word2Vec model was trained on the physician notes corpus with optimized parameter configuration generates effective word embedding features to be fed into the neural network model. The hybrid combination of these with the embedding features of the corpus generated by the pre-trained model trained on PubMed and PMC articles further enhanced and enriched the semantics of the textual features with biomedical domain knowledge. {\color{black}This is clear from the baseline comparison shown in Table \ref{results_compare} and it is this combination that has further enabled the FFNN to generalize better and learn the textual feature representation, effectively improving prediction performance when compared to the state-of-the-art model built on structured data. }

It is interesting to note that the patient data was modeled using only textual features, without any EHRs, structured data or other processed information. The high AUPRC and AUROC values obtained in comparison to the state-of-the-art's (based on structured data) performance is an indication that the unstructured text clinical notes (physician notes in this case) contain abundant patient-specific information that was beneficial for predictive analytics applications. Moreover, the conversion process from unstructured patient text reports to structured data can be eliminated, thereby saving huge man hours, cost and other resources. The proposed approach also eliminates any dependency on structured EHRs, thus making it suitable for deployment in developing countries. 

We found that the data preparation pipeline adopted for this study could be improved, as it sometimes resulted in some conflicting cases during training. This is because of the structure of the MIMIC-III dataset, in which, the physician notes do not have a direct assignment to ICD9 disease codes. To overcome this problem, we designed an assignment method in which we extracted ICD9 codes from the DIAGNOSES\_ICD table and assigned them to all patients with the same SUBJECT\_ID and HADM\_ID in the physician notes data subset. A side-effect of this method is that, at times, the ICD9 disease codes/groups assigned to the physician notes may not be related to that particular disease as a patient under same admission can have multiple physician notes. However, the model achieved promising results and the good performance indicates that prediction model was able to capture disease-specific features using the information present in the notes during the admission. 

\section{Conclusion and Future Work}
\label{conclusion}
In this paper, a deep neural network based model for predicting ICD9 disease groups from physician notes in the form of unstructured text is discussed. The proposed approach is built on a hybrid feature set consisting of word embedding features generated from a Word2Vec Skipgram model and also from a pre-trained word embedding model trained on biomedical articles from PubMed and PMC. The ICD9 disease codes were categorized into 20 standard groups and then used to train a binary classifier for multi-label prediction. The two sets of word embedding features were input into a neural network model as inputs and the weights of hybrid addition was determined dynamically using backpropagation. The combined features were further fed into a FFNN architecture to train the classifier and the prediction model was validated and benchmarked against state-of-the-art ICD9 disease group prediction model. The experiments highlighted the promising results achieved by the proposed model, outperforming the state-of-the-art model by 15\% in terms of AUROC and 40\% in terms of AUPRC. 

The promising results achieved by the proposed model underscores its usefulness as an alternative to current CDSS approaches, as it eliminates the dependency on structured clinical data, ensuring that hospitals in developing countries with low structured EHR adoption rate can also make use of effective CDSS in their functioning. As part of future work, we plan to address the issues observed in the current data preparation strategy, and enhance it by sorting out the disease group assignment problems. We also intend to explore other techniques to further optimize topic and feature modeling of textual representations to study their effect on disease prediction. 

\section*{Acknowledgement}
We gratefully acknowledge the use of the facilities at the Department of Information Technology, NITK Surathkal, funded by Govt. of India's DST-SERB Early Career Research Grant (ECR/2017/001056) to the second author.

%
%
%
 \let\clearpage\relax
 \vspace{1cm}
 \renewcommand{\bibname}{References}
 \bibliographystyle{splncsnat}
 \bibliography{ref}
 \newpage
\end{document}